\documentclass[journal]{IEEEtran}

\usepackage{graphicx}
\usepackage{color}
\usepackage{cite}
\usepackage{url}
\usepackage{subfigure}
\usepackage{bm}
\usepackage{amsmath}

\usepackage{subfigure}

\ifCLASSINFOpdf
\else
\fi

\begin{document}
\bstctlcite{IEEEexample:BSTcontrol}

\title{Watch from sky: machine-learning-based multi-UAV network for predictive police surveillance}

\author{Ryusei~Sugano,
 Ryoichi~Shinkuma,~\IEEEmembership{Senior member,~IEEE,}
 Takayuki~Nishio,~\IEEEmembership{Senior member,~IEEE,}
 Sohei~Itahara,~\IEEEmembership{Student member,~IEEE,}
 and~Narayan~B.~Mandayam,~\IEEEmembership{Fellow,~IEEE}

 \thanks{R. Sugano and R. Shinkuma are with the Faculty of Engineering, Shibaura Institute of Technology, 3-7-5 Toyosu, Koto-ku, Tokyo, 135-8548, Japan. e-mail: \{al19042, shinkuma\}@shibaura-it.ac.jp}
 \thanks{T. Nishio is with the School of Engineering, Tokyo Institute of Technology, Japan. e-mail:nishio@ict.e.titech.ac.jp}
 \thanks{S. Itahara is with the Graduate School of Informatics, Kyoto University, Kyoto 606-8501, Japan.}
 \thanks{N. B. Mandayam is with WINLAB, Department of Electrical and Computer Engineering, Rutgers University, US, narayan@winlab.rutgers.edu}
 \thanks{This work was supported in part by JST PRESTO Grant Numbers JPMJPR1854 and JPMJPR2035, and JSPS KAKENHI, Japan, under Grant Number 21H03427.}
}


\maketitle

\begin{abstract}
This paper presents the watch-from-sky framework, where multiple unmanned aerial vehicles (UAVs) play four roles, i.e., sensing, data forwarding, computing, and patrolling, for predictive police surveillance. Our framework is promising for crime deterrence because UAVs are useful for collecting and distributing data and have high mobility. Our framework relies on machine learning (ML) technology for controlling and dispatching UAVs and predicting crimes. This paper compares the conceptual model of our framework against the literature. It also reports a simulation of UAV dispatching using reinforcement learning and distributed ML inference over a lossy UAV network.
\end{abstract}


\IEEEpeerreviewmaketitle

\section{Introduction}
\label{sec:intro}
Reduction in police and security forces is an emerging problem in many parts of the world, leading to diminished public safety and increased crime. For example, in England and Wales, the number of police officers in 2016 was reported to be about 120,000, which is approximately 14\% less than in 2009.

Cloud enabled infrastructures with unmanned aerial vehicles (UAVs) and machine learning (ML) are emerging as promising approaches to improving patrolling capabilities \cite{miyano2020multi}. Studies have suggested that UAVs equipped with image sensors can be used for patrolling areas in the fight against crime. Such UAV policing systems have been reported to be effective in deterring a wide range of crimes, not just those that occur outside buildings. If police officers are patrolling an area, potential criminals therein will think their chances of making an escape are lower; this means that crimes inside and outside buildings in that area are less likely to occur. A UAV policing system tested in Mexico has been shown to have this effect.

The development of ML technology has made it possible to predict crime from a variety of crime-related data. Several states in the United States have adopted predictive police programs. In Japan, the Kanagawa Prefectural Police performed trials on predictive policing in anticipation of the Tokyo Olympics. Moreover, several law enforcement agencies, including Chicago Police, NYPD, and Boston Police, publish data relating crimes to the areas in which they occur on open data platforms such as Kaggle and the IBM Open Crime Data API, making it possible to better allocate personnel in advance to areas where crime is likely to occur.

This paper proposes the watch-from-sky framework, where multiple UAVs play four roles, i.e., sensing, data forwarding, computing, and patrolling, for predictive police surveillance. This framework leverages the mobility of UAVs with a cloud-enhanced infrastructure and ML in order to collect and distribute data to improve patrols to deter crime. A data-driven approach is taken where crime prediction data allows UAVs to be strategically distributed in more crime prone areas. Specifically, we use reinforcement learning to dispatch (steer) UAVs to geographical areas in a manner that improves both data acquisition and UAV utilization for improved crime prediction and deterrence.

The reinforcement-learning model leads the multi-UAV system to the optimal solution for task allocation and placement of UAVs while considering both data collection for crime prediction and UAV placement for crime deterrence. The task allocation and placement enables areas where crimes will likely occur to be covered efficiently, thereby improving UAVs' crime deterrence capability. This paper also presents a packet-loss resilient distributed inference (DI) that enables UAVs to conduct ML inference cooperatively in lossy wireless networks. By tuning the model with the dropout technique, the DI method improves the tolerance to missing data induced by packet loss of a deep neural network for locally predicting crimes. With this model, the UAVs can conduct DI even in lossy networks.

\section{Proposed system}
\subsection{System overview}
\begin{figure}[!t]
 \centering
 \includegraphics[width=\linewidth]{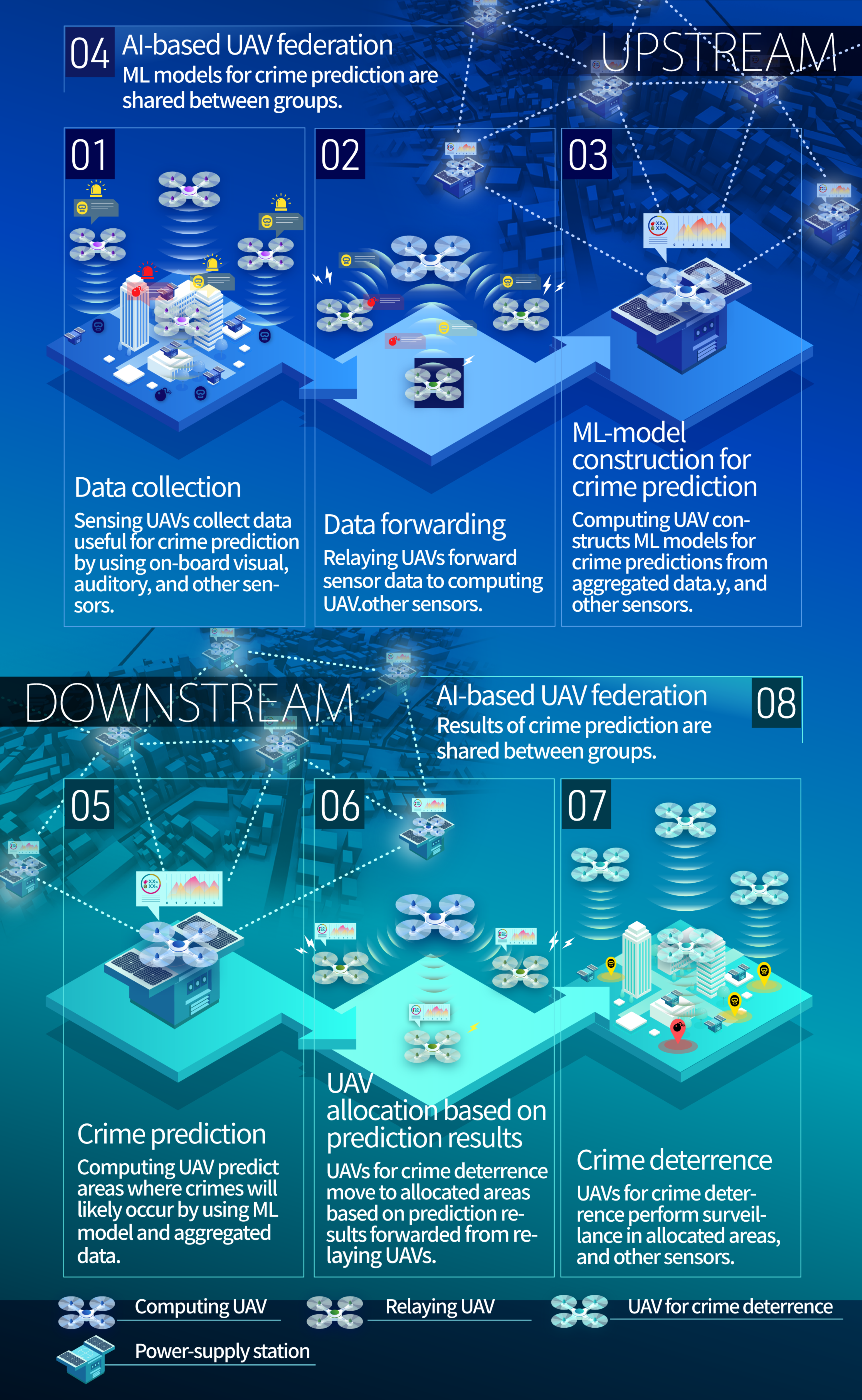}
 \caption{Conceptual illustration of watch-from-sky framework}
 \label{fig:wfs}
\end{figure}

Fig.~\ref{fig:wfs} shows a conceptual illustration of our watch-from-sky framework. UAVs adaptively switch between sensing, data forwarding, computing, or patrolling for crime deterrence when in the air and charge their batteries when they go back to the power supply station they belong to. In the upstream case, sensing UAVs first collect useful data for making crime predictions by using on-board visual, auditory, and other sensors. By `useful' data we mean data showing correlations with crimes and from which ML can predict crimes. The collected sensor data are forwarded by relaying UAVs to the computing UAV. The computing UAV constructs ML models for crime prediction from the data. The constructed models are shared between groups that consists of sensing, relaying, and computing vehicles; this is called AI-based UAV federation in our framework. In the downstream case, the computing UAV in each group predicts areas where crimes will likely occur by using the ML model and the collected sensor data. Predictions are shared by the group. The UAVs for crime deterrence move to their allocated areas based on the predictions forwarded by the relaying UAVs. Finally, the UAVs for crime deterrence perform police surveillance in the allocated areas.

\subsection{Related work}
\label{sec:related}
\subsubsection{UAV networks for predictive police surveillance}
Jin~et al. \cite{jin2020uav}designed optimization algorithms and scheduling strategies for UAV clusters. They considered how to dispatch UAVs in response to a video surveillance event, how to improve the dispatch efficiency and the video data processing efficiency of UAV clusters, how to balance the flight efficiency of UAVs with the response efficiency to video events, and how to allocate UAVs, radio base stations, and video surveillance devices.

Yan et al. investigated issues when UAVs and police vehicles on the ground cooperate \cite{yang2021programming}. Faced with the uncertainty of the patrol environment and patrol resources, their model guarantees the deterrence and emergency response capabilities of patrol missions by optimizing the allocation of patrol points and patrol routes.

Trotta~et al. presented a network architecture and a supportive optimization framework that enable UAVs to perform city-scale video surveillance of a range of points of interest (PoI), such as tourist attractions \cite{trotta2018uavs}. They assumed that the UAVs can land on public transport buses and ``ride'' on them to the selected PoI as they recharge their batteries.

Miyano~et al. presented a comprehensive multi-UAV allocation framework for predictive crime deterrence and data acquisition that works with predictive models using the ML approach \cite{miyano2020multi}. Their framework determines the most effective placement of UAVs to maximize the chances of arresting criminals, while at the same time acquiring data that helps to improve subsequent crime predictions.

Gassara and Rodriguez developed an architectural model of a distributed system to support UAV group collaboration in the context of search and rescue missions \cite{gassara2021describing}. They focused on modeling adaptative cooperation to maintain mission requirements while meeting environmental and resource constraints. In search and rescue missions, the processing time and the data transfer time of the acquired data are both important. Miyano~et al. presented a scheduling method for a multi-UAV search system that takes into account both the image-data processing time and data transfer time \cite{miyano2019utility}.

\subsubsection{UAV allocation and reinforcement learning}
The authors presented the design of a wireless mesh network that relies on UAV's automation (AP separation and battery replacement) capabilities \cite{shinkuma2020design}. This design includes a mathematical formulation of the network models and the UAV scheduling algorithm. They took a heuristic approach, though it was reported that reinforcement learning could be a solution to such a UAV allocation problem. Liu~et al. presented a two-level quasi-distributed control framework for UAVs for persistent ground surveillance in unknown urban areas \cite{liu2020reinforcement}. In their framework, targets are specified through high-level control strategies for cooperative surveillance operations, while trained artificial neural networks are responsible for low-level UAV maneuvering controls for target homing and collision avoidance.

\subsection{DI in resource constrained networks}
The DI framework has been studied as a way to enable inference with cutting-edge deep learning on computationally poor networked devices (e.g., UAVs, IoT devices, and connected vehicles), \cite{Zhang20, Mohammed20, Shao20}. In this framework, all or part of a computationally expensive task is offloaded from the UAVs to other UAVs or edge servers to reduce computation latency. Zhang et al.\ introduced an edge intelligence system for intelligent internet of vehicle (IoV) environments including edge computing and edge AI \cite{Zhang20}.

Mohammed~et al.\ proposed a method to adaptively divide a deep neural network (DNN) into multiple portions and offload computations on the basis of a matching game \cite{Mohammed20}. They evaluated their method on a self-driving car dataset and showed that it significantly reduces the total latency of inference.

Shao et al. presented a device-edge co-inference framework for resource-constrained devices \cite{Shao20}. In this framework, a model is split at the optimal point, and the on-device computation and resulting communication overhead are reduced by using communication-aware model compression. The communication overhead is further reduced by using task-oriented encoding of the intermediate features.

\section{Joint optimization of UAV placements and roles using reinforcement learning}
\subsection{System model}
\begin{figure}[!t]
 \centering
 \includegraphics[width=\linewidth]{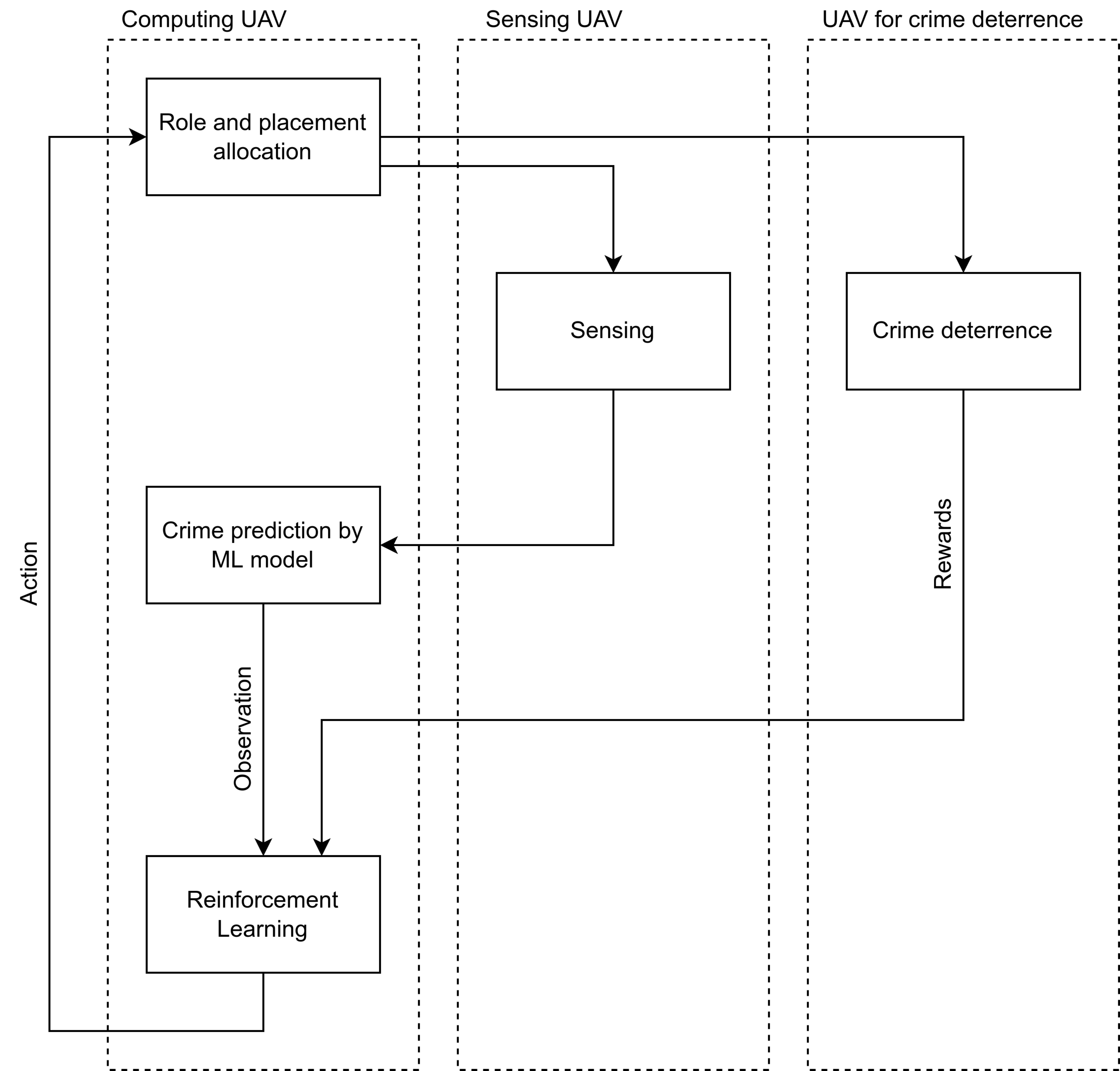}
 \caption{Evaluation model for task and placement optimization using reinforcement learning.}
 \label{fig:3_system_model}
\end{figure}

This section describes the system model for the RL-based joint optimization of placements and roles of UAVs, as shown in Fig.~\ref{fig:3_system_model}. The system consists of a power-supply station and multiple UAVs with three roles: sensing, computing, and deterrence. The UAVs move from the power-supply station to their respective locations and return to the power-supply station when their battery level is low. The sensing UAV senses a specific range of the placed area with its onboard sensors and obtains valuable information with which to predict crimes, including brightness, human flow, weather, and temperature. It also relays sensing data to the computing UAV by connecting to other sensing UAVs and computing UAVs within its communication range. The computing UAV uses the data collected from the connected sensing UAVs to predict where serious crimes are likely to occur and uses this information to make the UAV placements in the next round. The UAVs for crime deterrence aid police patrols in preventing crimes by taking videos and sounding alerts to deter criminals within a specific geographical area covered by the UAV. Given this model, we formulate the problem as follows.
\begin{quote}
Objective:

Maximize the number of crimes deterred by the UAVs divided by the number of potential crime offenses,

Constraint:

The total number of computing UAVs, sensing UAVs, and UAVs for crime deterrence is smaller than or equal to the total number of available UAVs.
\end{quote}
The objective of the joint optimization is to maximize the number of crimes deterred by the UAVs. By increasing the coverage of the sensing UAVs and making more data available to the computing UAVs (i.e., increasing the number of sensing and computing UAVs and placing them in appropriate areas), the prediction accuracy increases, and thereby, the UAVs for crime deterrence can be efficiently placed. However, increasing the number of sensing and computing UAVs decreases the number of UAVs for crime deterrence as there would be a limit to the total number of UAVs; this degrades the maximum coverage of the UAVs for crime deterrence.

\subsection{Methodology}
\label{sec:methodology}
Simply placing UAVs for crime deterrence randomly is not enough to deter crimes in a wide area. In the proposed method, the computing UAV uses the data collected by the sensing UAV and a ML model to predict the number of crimes per block that will occur in the next period. In accordance with the predictions, the UAVs are assigned to different locations and roles for the next period. Reinforcement learning is used to optimize the placement and role assignment of UAVs so as to maximize the number of crimes deterred. Specifically, the location and role assignment of each UAV are learned as action spaces, the number of crimes per block predicted by the computing UAV as observation spaces, and the number of crimes deterred as the number of rewards.

\subsection{Evaluation}
\label{sec:evaluation}
We evaluated the effectiveness of the proposed system in a simulation using a real crime dataset. As a comparison, we also simulated the case where all UAVs are for crime deterrence and placed at random. We trained the system using training data and evaluated it on test data. In the evaluation, the system follows the procedure described in the previous section.

\subsubsection{Simulation parameters}
This section describes the parameters of the simulation. We used the crime dataset \cite{crimes-in-chicago} available on Kaggle. Specifically, we used the data from 7:00 pm to 12:00 pm on Fridays, when crime is particularly high. The dataset can be divided into 25 regions for the whole area, which also can be divided into 303 blocks for the whole area, and we used 12 blocks of Region 6, which is one of the regions with a high number of crimes. We used data from 2005 to 2013 as the training data and data from 2014 to 2016 as the test data. Since the frequency of crime occurrence for a block is not high, to increase the frequency of data for training, we aggregated data from 2014 to 2016 into one year; we dealt crime occurrences at the same time, on the same day, in the same week, and in the same month but in different years as the ones in the same year. The major crimes were robbery, sexual assault, murder, and arson, and the other crimes were misdemeanors. Crimes that were deterred were considered major crimes, and information obtained through sensing was considered misdemeanors. Note that Miyano et al. showed that there is a correlation between misdemeanors and serious crimes \cite{miyano2020multi}.

The control cycle started at 7 pm and ended at 12 pm. The power-supply station was at the police station, and the power supply was simulated by the fact that each UAV could only be placed within a certain range from the power-supply station to which it belongs. The total number of UAVs is 20. The communication range of relay UAVs was 500 m, the sensing range of sensing UAVs was 100 m, and the crime deterrent range of the UAVs for crime deterrence was 80, 160, 320, 640, and 1280 m. In addition, to improve the efficiency of reinforcement learning, each UAV could be placed in accordance with a grid of 50 m width in advance. We assumed that the computing UAVs shared the crime prediction data with each other and trained one identical model for the whole.

\subsubsection{Reinforcement learning algorithm}
This section describes the reinforcement learning method used to optimize the placement and role of each UAV. The reinforcement learning algorithm was proximal policy optimization algorithms (PPO) \cite{schulman2017proximal}. We used python and its libraries for the implementation. The environment was created using the OpenAI Gym and the PPO was Stable Baselines3. We set the number of training steps to 1e7 and terminated in the middle when the reward became stable. The other parameters were set to the default values.

\subsubsection{Crime estimation algorithm}
This section describes the ML algorithm used to predict the number of crimes per block for the next control cycle on the basis of the sensed data of the connected UAVs. LSTM was used for making the prediction. Misdemeanor crime data acquired by the sensing UAV connected to the computing UAV were counted in each block and entered in the number of blocks’ dimension. Blocks where no crime was committed were given a value of 0. The LSTM model we used had 100 LSTM units in the hidden layers, which were fully connected to the dense layers. ReLU was used as the activation function. The output was the number of crimes per block for the next control cycle. The model was trained using the Adam optimizer in 100 epochs and with a batch size of 100.

\subsubsection{Results}
\label{sec:results}
\begin{figure}[h]
 \centering
 \includegraphics[width=\linewidth]{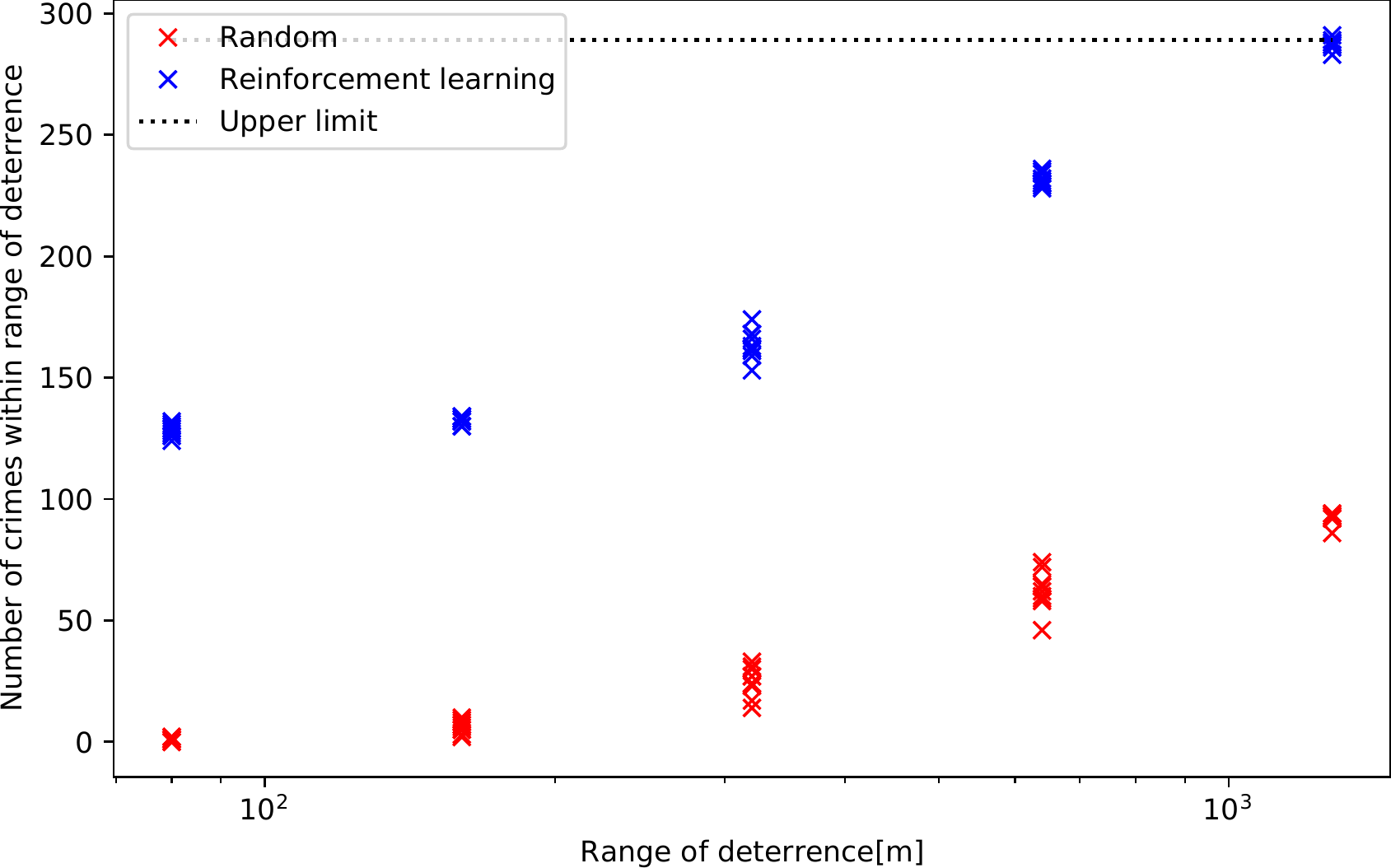}
 \caption{Number of crimes within range of deterrence vs. range of deterrence.}
 \label{fig:3_results}
\end{figure}

Fig.~\ref{fig:3_results} shows the number of crimes that can be deterred by UAVs for crime deterrence while varying the range of deterrence of the UAV for the cases of reinforcement learning and testing on the test data. Ten tests were conducted at each possible crime range of deterrence, and the average is shown as a line. The horizontal axis indicates the crime deterrence distance on a log scale, while the vertical axis indicates the number of deterred crimes. When all UAVs were allocated the role of crime deterrence and placed randomly, the UAVs could barely predict crimes when the crime range of deterrence was 80 m. Even when the crime range of deterrence was 1280 m, they could only deter about 30\% of the crimes. When the placement and role of each UAV were optimized using reinforcement learning, a little less than half of the crimes were deterred when the deterrence distance was 80 m, whereas almost all of the crimes were deterred when the deterrence distance was 1280 m. In all cases, the use of reinforcement learning improved performance by more than 40\% compared with placement of UAVs for crime deterrence alone. These results show that reinforcement learning is effective in optimizing the placements and roles of UAVs.

\section{DI over lossy UAV networks}
This section discusses how to perform ML inference on UAV networks, where UAVs generally have poor computing power and the wireless links among UAVs are lossy. As described in the previous section, ML inference plays an important role in predicting crimes and determining optimal UAV placement. In addition, computer vision using state-of-the-art ML makes it possible to detect real-world anomalies from camera images \cite{Sultani_2018_CVPR}, thereby enabling prediction and prevention of crime and violence at the edge. However, a question arises as to how to perform such ML inferences with UAVs that lack computational power in lossy wireless networks.

We have proposed a packet-loss resilient DI that enables UAVs to conduct ML inference cooperatively with lossy wireless connections \cite{itahara21}. This method improves the ML model's tolerance to missing data, which eliminates the need for packet retransmissions to compensate for packet loss in the wireless link and thereby reduces traffic and communication latency induced by packet loss in lossy networks.

\subsection{System model}
\begin{figure}[t!]
 \centering
 \includegraphics[width=\linewidth]{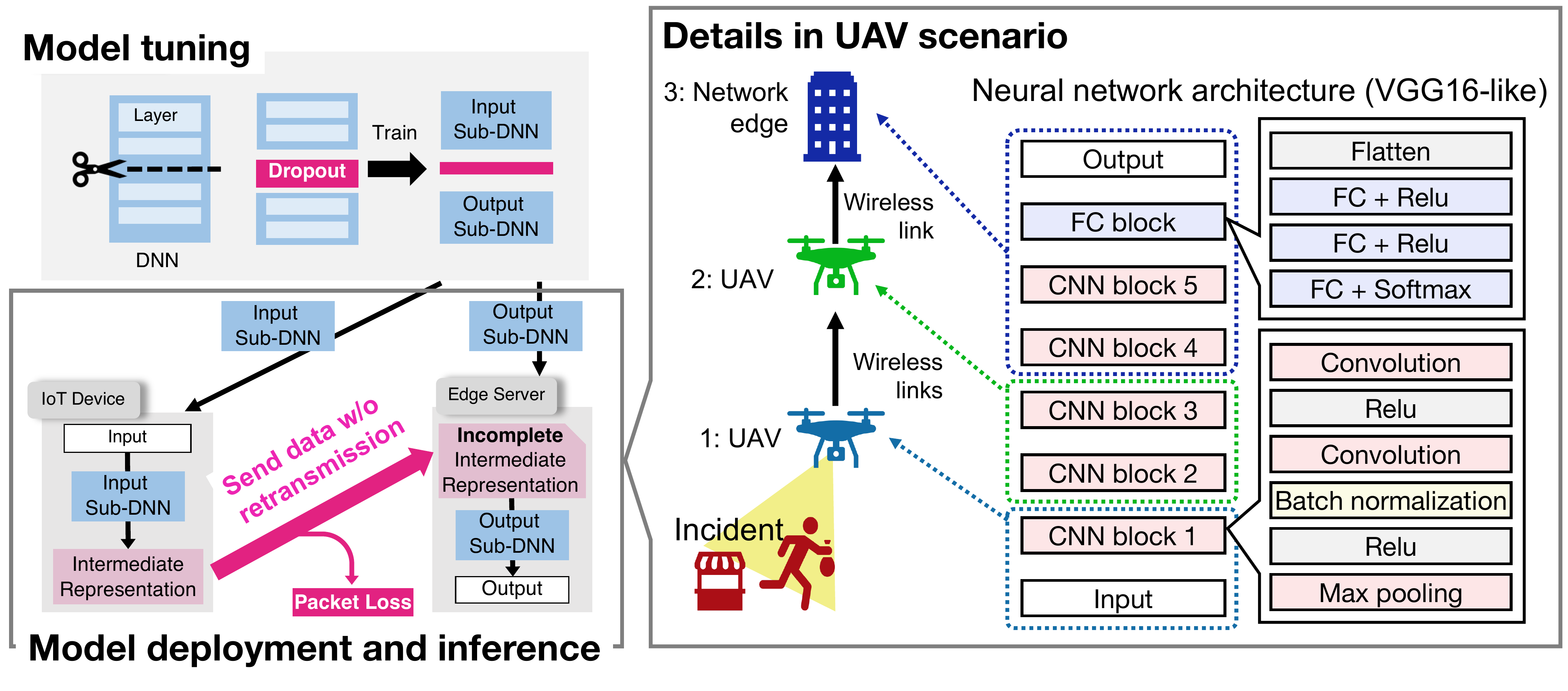}
 \caption{Packet-loss resilient DI. Model training with the dropout technique improves tolerance to missing data, thereby enabling low-latency but highly accurate inference without retransmissions in a lossy network.}
 \label{fig:4_DI}
\end{figure}
We assume that UAVs equipped with cameras are connected to adjacent UAVs to form a multi-hop wireless network. Packet loss occurs probabilistically in the wireless communication link due to congestion and interference. UAVs cooperatively conduct crime predictions from their camera imagery with a well-trained NN model via the UAV network. The trained model is separated into portions, called sub-NNs, and the sub-NNs are deployed in the UAVs. The model is separated into three parts: input-subNN, middle-subNN, and output-subNN. When inference is conducted, an image is inputted to the input-subNN and the output of the input-subNN is forwarded to middle-subNN. The output of middle-subNN is forwarded to the output-subNN, and the inference result is obtained from the output-subNN.

\subsection{Methodology}
The challenge of this study is enabling DI in unreliable UAV networks without degrading the accuracy or increasing communication overheads (e.g., retransmissions and rate control). To this end, our key idea is to train DNN by emulating the effect of packet drops by using a dropout technique that randomly drops the activations in the DNN. Dropout was initially proposed as an ML technique to improve model performance with a regularization effect. We utilized it to simulate packet drops in lossy UAV networks in the model training. The model trained with dropout could make accurate predictions from messages corrupted by packet loss in the networks. The reader may refer to \cite{itahara21} for the detailed training procedure.

\subsection{Evaluation}
We evaluated our method on an image classification task, CIFAR-10. We used a convolutional neural network (CNN) consisting of five convolutional blocks and a fully connected (FC) block, which was designed with reference to VGG16, as shown in Fig.~\ref{fig:4_DI}. The model was trained with the CIFAR-10 training dataset, which has 50,000 images. In the proposed method, dropout layers with dropout rates $r = \{0.1, 0.3, 0.5\}$ were inserted at the end of each CNN block, and the model was fine-tuned with the same training dataset. The conventional method used the model without fine-tuning. We assumed that an inference task is offloaded to three UAVs; namely, the model was separated into three portions and deployed to three UAVs. The UAVs were connected in a chain via a wireless link with a packet loss rate $p_{i,j}$, where $i$ and $j$ identify the UAVs. We assumed that no retransmissions occurred in the transport or the MAC layer. Therefore, $p_{i,j}$ of the intermediate representations was randomly lost in each wireless link.

\begin{figure}[t!]
\centering
 \subfigure[Model was split at CNN blocks 1 and 3]{
 \includegraphics[width=0.8\linewidth]{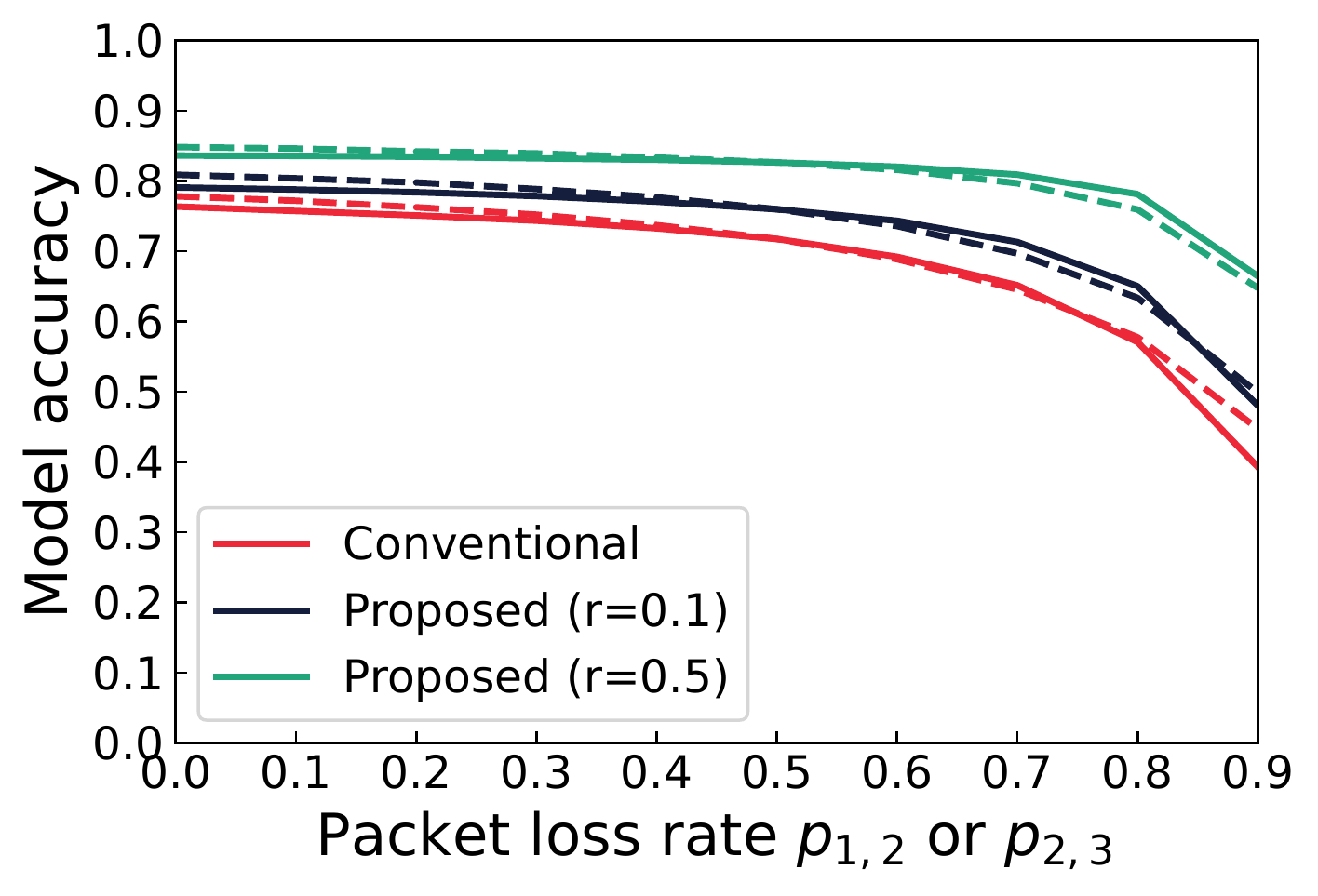}
 \label{fig:4_cnn3}
 }

 \subfigure[Model was split at CNN block 1 and 4]{
 \includegraphics[width=0.8\linewidth]{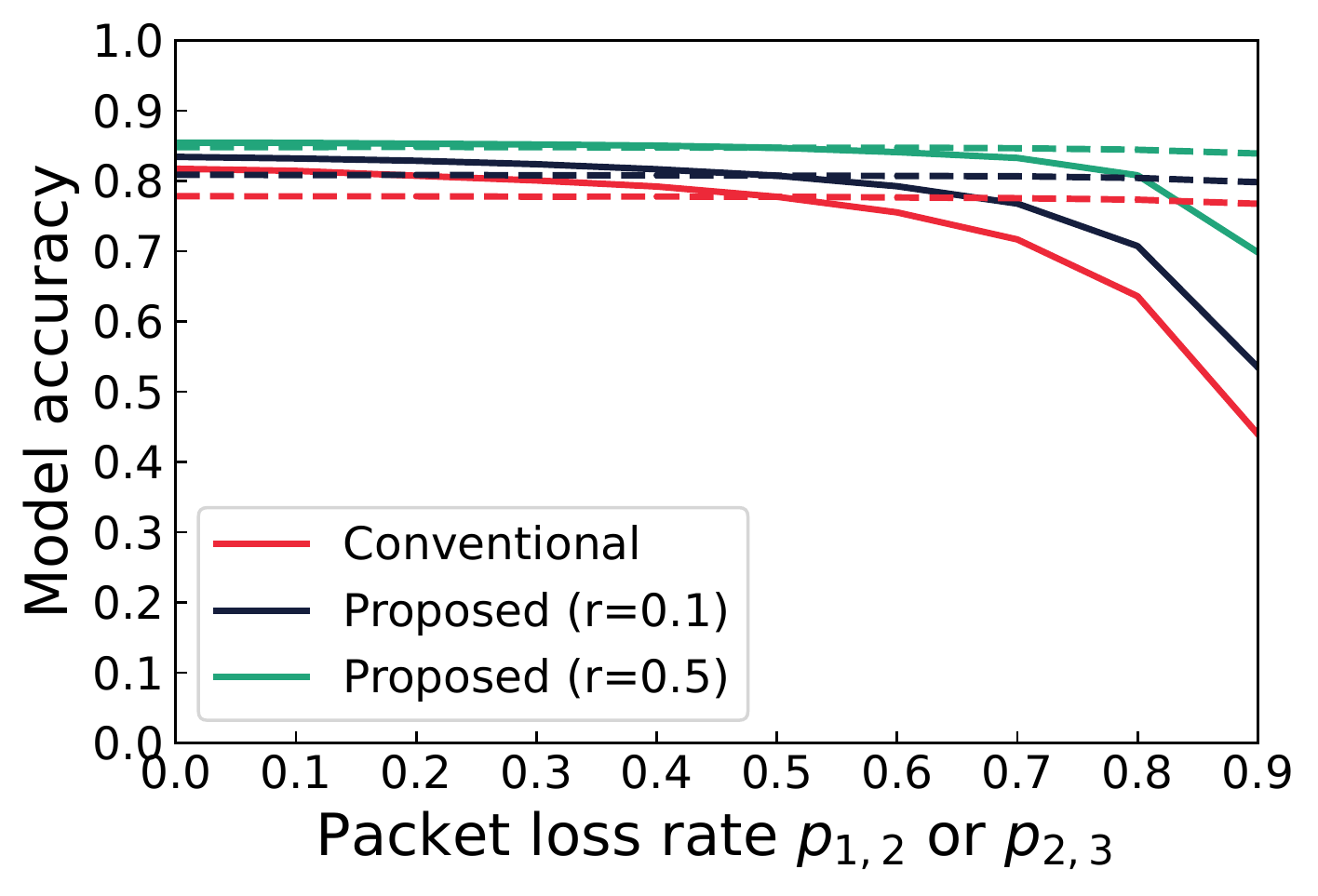}
 \label{fig:4_cnn4}
 }
 \caption{Model accuracy as a function of packet loss rate. The solid lines indicate the results when $p_{1,2}$ was changed and $p_{2,3}$ was fixed at 0.5, and the dashed lines indicate the results when $p_{1,2}$ was fixed at 0.5 and $p_{2,3}$ was changed. }
 \label{fig:4_results}
\end{figure}

Fig.~\ref{fig:4_cnn3} plots model accuracy as a function of packet loss rate of the wireless links $p_{1,2}$ and $p_{2,3}$ when the model was split at CNN block 1 and 3. For all methods, the model accuracy decreased as the packer loss rate$p_{1,2}$ and $p_{2,3}$ increased. However, the proposed method was more accurate than the conventional method and maintained accuracy, with a decrease of less than 5 points up to a packet loss ratio of 0.8. These results show that the proposed method enables DI without retransmissions in a UAV network with high packet loss. Fig.~\ref{fig:4_cnn4} plots model accuracy as a function of the packet loss rate $p_{1,2}$ and $p_{2,3}$ when the model was split at CNN block 1 and 4. As shown by the dashed lines in Fig.~\ref{fig:4_cnn4}, even as the packet loss rate $p_{2,3}$ increased, the accuracy remained about the same for all methods, which indicates
that high packet-loss tolerance can also be achieved by optimizing the partitioning position of the model.
These results demonstrate the feasibility of DI that achieves both high prediction accuracy and low communication overhead in a lossy UAV network.

\section{Conclusion}
\label{sec:conclusion}
This paper proposed the watch-from-sky framework, where multiple UAVs adaptively work on four predictive police surveillance tasks: sensing, data forwarding, computing, and patrolling. In our framework, UAVs work together to collect and distribute data and patrol for crime deterrence. We studied the issue of task allocation and placement of multiple UAVs using reinforcement learning. In particular, we conducted a performance evaluation in terms of the number of crimes deterred by UAVs to determine a proper task allocation and placement that would enable areas where crimes likely occur to be covered efficiently. We also studied the issue of developing a packet-loss resilient DI that enables UAVs to conduct ML inference cooperatively in lossy wireless networks. The results of our study showed that, even in lossy networks, the UAVs can conduct DI with the model without having to use retransmission or a low transmission rate.

\ifCLASSOPTIONcaptionsoff
 \newpage
\fi

\bibliographystyle{IEEEtran}
\bibliography{reference_uav_rl2}

\begin{thebibliography}{10}
\providecommand{\url}[1]{#1}
\csname url@samestyle\endcsname
\providecommand{\newblock}{\relax}
\providecommand{\bibinfo}[2]{#2}
\providecommand{\BIBentrySTDinterwordspacing}{\spaceskip=0pt\relax}
\providecommand{\BIBentryALTinterwordstretchfactor}{4}
\providecommand{\BIBentryALTinterwordspacing}{\spaceskip=\fontdimen2\font plus
\BIBentryALTinterwordstretchfactor\fontdimen3\font minus
  \fontdimen4\font\relax}
\providecommand{\BIBforeignlanguage}[2]{{%
\expandafter\ifx\csname l@#1\endcsname\relax
\typeout{** WARNING: IEEEtran.bst: No hyphenation pattern has been}%
\typeout{** loaded for the language `#1'. Using the pattern for}%
\typeout{** the default language instead.}%
\else
\language=\csname l@#1\endcsname
\fi
#2}}
\providecommand{\BIBdecl}{\relax}
\BIBdecl

\bibitem{miyano2020multi}
K.~Miyano, R.~Shinkuma, N.~Shiode, S.~Shiode, T.~Sato, and E.~Oki, ``Multi-uav
  allocation framework for predictive crime deterrence and data acquisition,''
  \emph{Internet of Things}, vol.~11, p. 100205, 2020.

\bibitem{jin2020uav}
Y.~Jin, Z.~Qian, and W.~Yang, ``Uav cluster-based video surveillance system
  optimization in heterogeneous communication of smart cities,'' \emph{IEEE
  Access}, vol.~8, pp. 55\,654--55\,664, 2020.

\bibitem{yang2021programming}
J.~Yang, Z.~Ding, and L.~Wang, ``The programming model of air-ground
  cooperative patrol between multi-uav and police car,'' \emph{IEEE Access},
  vol.~9, pp. 134\,503--134\,517, 2021.

\bibitem{trotta2018uavs}
A.~Trotta, F.~D. Andreagiovanni, M.~Di~Felice, E.~Natalizio, and K.~R.
  Chowdhury, ``When uavs ride a bus: Towards energy-efficient city-scale video
  surveillance,'' in \emph{Proc. IEEE International Conference on Computer
  Communications (INFOCOM)}.\hskip 1em plus 0.5em minus 0.4em\relax IEEE, 2018,
  pp. 1043--1051.

\bibitem{gassara2021describing}
A.~Gassara and I.~B. Rodriguez, ``Describing correct uavs cooperation
  architectures applied on an anti-terrorism scenario,'' \emph{Journal of
  Information Security and Applications}, vol.~58, p. 102775, 2021.

\bibitem{miyano2019utility}
K.~Miyano, R.~Shinkuma, N.~B. Mandayam, T.~Sato, and E.~Oki, ``Utility based
  scheduling for multi-uav search systems in disaster-hit areas,'' \emph{IEEE
  Access}, vol.~7, pp. 26\,810--26\,820, 2019.

\bibitem{shinkuma2020design}
R.~Shinkuma and N.~B. Mandayam, ``Design of ad hoc wireless mesh networks
  formed by unmanned aerial vehicles with advanced mechanical automation,'' in
  \emph{Proc. IEEE International Conference on Distributed Computing in Sensor
  Systems (DCOSS)}.\hskip 1em plus 0.5em minus 0.4em\relax IEEE, 2020, pp.
  288--295.

\bibitem{liu2020reinforcement}
Y.~Liu, H.~Liu, Y.~Tian, and C.~Sun, ``Reinforcement learning based two-level
  control framework of uav swarm for cooperative persistent surveillance in an
  unknown urban area,'' \emph{Aerospace Science and Technology}, vol.~98, p.
  105671, 2020.

\bibitem{Zhang20}
J.~Zhang and K.~B. Letaief, ``Mobile edge intelligence and computing for the
  internet of vehicles,'' \emph{Proceedings of the IEEE}, vol. 108, no.~2, pp.
  246--261, 2020.

\bibitem{Mohammed20}
T.~Mohammed, C.~Joe-Wong, R.~Babbar, and M.~D. Francesco, ``Distributed
  inference acceleration with adaptive dnn partitioning and offloading,'' in
  \emph{Proc. IEEE International Conference on Computer Communications
  (INFOCOM)}, July 2020, pp. 854--863.

\bibitem{Shao20}
J.~Shao and J.~Zhang, ``Communication-computation trade-off in
  resource-constrained edge inference,'' \emph{IEEE Communications Magazine},
  vol.~58, no.~12, pp. 20--26, 2020.

\bibitem{crimes-in-chicago}
Kaggle, ``Crimes in chicago,''
  \url{https://www.kaggle.com/currie32/crimes-in-chicago}, 2018, (Accessed on
  Dec 1, 2021).

\bibitem{schulman2017proximal}
J.~Schulman, F.~Wolski, P.~Dhariwal, A.~Radford, and O.~Klimov, ``Proximal
  policy optimization algorithms,'' \emph{arXiv preprint arXiv:1707.06347},
  2017.

\bibitem{Sultani_2018_CVPR}
W.~Sultani, C.~Chen, and M.~Shah, ``Real-world anomaly detection in
  surveillance videos,'' in \emph{Proc. IEEE Conference on Computer Vision and
  Pattern Recognition (CVPR)}, June 2018, pp. 6479--6488.

\bibitem{itahara21}
S.~Itahara, T.~Nishio, and K.~Yamamoto, ``Packet-loss-tolerant split inference
  for delay-sensitive deep learning in lossy wireless networks,'' in
  \emph{Proc. IEEE Globecom}, Dec. 2021, pp. 1--6.

\end{thebibliography}

\begin{IEEEbiography}
 Ryusei Sugano is a student in the Faculty of Engineering, Shibaura Institute of Technology, Japan. His research interests include the design of social information network systems.
\end{IEEEbiography}

\begin{IEEEbiography}
 Ryoichi Shinkuma received the Ph.D. degree from Osaka University in 2003. He was an assistant/associate professor at Kyoto University until 2021. He was a visiting scholar at WINLAB, Rutgers University from 2008 to 2009. He is currently a professor in the Faculty of Engineering, Shibaura Institute of Technology.
\end{IEEEbiography}

\begin{IEEEbiography}
 Takayuki Nishio is an associate professor at the School of Engineering, Tokyo Institute of Technology, Japan. He received his B.E. degree in electrical and electronic engineering, and his Master's and Ph.D. degrees in informatics from Kyoto University in 2010, 2012, and 2013, respectively. From 2016 to 2017, he was a visiting researcher at WINLAB, Rutgers University, New Jersey. His current research interests include machine-learning-based network control, machine learning in wireless networks, and heterogeneous resource management.
\end{IEEEbiography}

\begin{IEEEbiography}
 Sohei Itahara received the B.E. degree in electrical and electronic engineering from Kyoto University in 2020. He is currently studying toward the M.I. degree at the Graduate School of Informatics, Kyoto University. He is a student member of the IEEE.
\end{IEEEbiography}

\begin{IEEEbiography}
 Narayan B. Mandayam is a Distinguished Professor and Chair of Electrical and Computer Engineering at Rutgers University, where he also serves as Associate Director of the WINLAB. His research contributions have been recognized with the 2015 IEEE Communications Society Advances in Communications Award for his work on power control and pricing, the 2014 IEEE Donald G. Fink Award for his IEEE Proceedings paper titled ``Frontiers of Wireless and Mobile Communications'' and the 2009 Fred W. Ellersick Prize from the IEEE Communications Society for his work on dynamic spectrum access models and spectrum policy. He is also a recipient of the Peter D. Cherasia Faculty Scholar Award from Rutgers University (2010), the National Science Foundation CAREER Award (1998), the Institute Silver Medal from the Indian Institute of Technology (1989) and its Distinguished Alumnus Award (2018). He is a Fellow and Distinguished Lecturer of the IEEE.
\end{IEEEbiography}

\end{document}